\pdfoutput=1

\documentclass[11pt,a4paper]{article}
\PassOptionsToPackage{hyphens}{url}\usepackage[hyperref]{emnlp2020}
\usepackage{times}
\usepackage{latexsym}
\usepackage{xspace}

\usepackage{microtype}

\usepackage[final]{changes}

\usepackage{subcaption}
\usepackage{graphicx} 
\graphicspath{{figures/}} 

\usepackage{tikz}
\usetikzlibrary{tikzmark}

\usepackage{multirow}

\usepackage{algorithmic}

\usepackage{amsmath}

\definechangesauthor[name=Mika Juuti, color=blue]{M}
\definechangesauthor[name=Tommi Gröndahl, color=red]{T}
\definechangesauthor[name=N. Asokan, color=brown]{A}
\aclfinalcopy 

\newcommand{\test}{{\textsc{Test}}\xspace}
\newcommand{\gold}{{\textsc{Gold Standard}}\xspace}
\newcommand{\seed}{{\textsc{Seed}}\xspace}
\newcommand{\simple}{\textsc{Copy}\xspace} 
\newcommand{\eda}{\textsc{EDA}\xspace} 
\newcommand{\add}{\textsc{Add}\xspace} 
\newcommand{\wn}{\textsc{WordNet}\xspace} 
\newcommand{\ppdb}{\textsc{PPDB}\xspace} 
\newcommand{\gensim}{\textsc{GloVe}\xspace} 
\newcommand{\bpemb}{\textsc{BPEmb}\xspace} 
\newcommand{\gpt}{\textsc{GPT-2}\xspace} 
\newcommand{\rocauc}{\textsc{ROC-AUC}\xspace}

\newcommand{\seedL}{\textsl{No Oversampling}\xspace} 
\newcommand{\simpleL}{\textsl{Simple Oversampling}\xspace} 
\newcommand{\edaL}{\textsl{Easy Data Augmentation}\xspace} 
\newcommand{\addL}{\textsl{Add Majority-class Sentence}\xspace} 
\newcommand{\wnL}{\textsl{Word Substitutions}\xspace} 
\newcommand{\ppdbL}{\textsl{Phrase Substitutions}\xspace} 
\newcommand{\gensimL}{\textsl{Word Substitutions}\xspace} 
\newcommand{\bpembL}{\textsl{Subword Substitutions}\xspace} 
\newcommand{\gptL}{\textsl{Conditional Generation}\xspace} 

\newcommand{\ab}{\textsc{AB}\xspace}
\newcommand{\abg}{\textsc{ABG}\xspace} 
\newcommand{\abgL}{\textsl{Mixed Augmentation} (\add, \bpemb \& \gpt)\xspace}

\newcommand{\wordlr}{\textsf{\small{Word-LR}}\xspace} 
\newcommand{\charlr}{\textsf{\small{Char-LR}}\xspace} 
\newcommand{\cnn}{\textsf{\small{CNN}}\xspace}
\newcommand{\bert}{\textsf{\small{BERT}}\xspace}

\newcommand{\numreps}{{30}\xspace}

\def\myemph#1{\emph{#1}}

\def\myparagraph#1{\noindent\textbf{#1}}
\def\myparagraphS#1{\noindent\textbf{#1}.}

\def\taa#1{\underline{\underline{\textbf{#1}}}}
\def\ta#1{{\textbf{#1}}}
\def\marg#1{\textbf{#1}}

\def\cls#1{\texttt{#1}}

\title{A little goes a long way: Improving toxic language classification despite data scarcity}

\author{Mika Juuti$^1$, Tommi Gr\"{o}ndahl$^2$, Adrian Flanagan$^3$, N. Asokan$^{1,2}$ \\
  University of Waterloo$^1$ \\ Aalto University$^2$ \\ Huawei Technologies Oy (Finland) Co Ltd$^3$ \\
  \texttt{mika.juuti@kela.fi, tommi.grondahl@aalto.fi} \\ \texttt{adrian.flanagan@huawei.com, asokan@acm.org}}

\date{}

\begin{document}
\maketitle

\begin{abstract}

Detection of some types of \myemph{toxic language} is hampered by extreme scarcity of labeled training data. \myemph{Data augmentation} -- generating new synthetic data from a labeled seed dataset --  can help. The efficacy of data augmentation on toxic language classification has not been fully explored. We present the first \myemph{systematic} study on how data augmentation techniques impact performance across toxic language classifiers, ranging from shallow logistic regression architectures to \added[id=T]{BERT -- a state-of-the-art pre-trained Transformer network.} We compare the performance of \myemph{eight techniques} on very scarce seed datasets. We show that while BERT performed the best,  shallow classifiers performed comparably when trained on data augmented with a combination of three techniques, including GPT-2-generated sentences. 
We discuss the interplay of performance and computational overhead, which can inform the choice of techniques under different constraints.
\end{abstract}

 \section{Introduction}
\label{sec:introduction}

\myemph{Toxic language} is an increasingly urgent challenge in online communities
~\cite{mathew-etal-2019}.
Although there are several datasets, most commonly from Twitter or forum discussions~\cite{Badjatiya2017, Davidsonetal2017, Waseemetal2016, Wulczynetal2017, Zhangetal2017},
high \myemph{class imbalance} is a problem with certain classes of toxic language \cite{breitfeller2019finding}. Manual labeling of toxic content is onerous, hazardous~\cite{facebook2020settlement}, and thus expensive.

One strategy for mitigating these problems is \myemph{data augmentation}~\cite{wang-yang-2015, ratner-etal-2017, wei-zou-2019}: complementing the manually labeled \myemph{seed data} with new synthetic documents.
The effectiveness of data augmentation for toxic language classification has not yet been thoroughly explored. 
On relatively small toxic language datasets, \emph{shallow classifiers} have been shown to perform well~\cite{Grondahletal2018}. 
At the same time, pre-trained \emph{Transformer} networks~\cite{vaswani2017attention} have led to impressive results in several NLP tasks~\cite{young2018recent}.
Comparing the effects of data augmentation between shallow classifiers and pre-trained Transformers is thus of particular interest.



We systematically \added[id=T]{compared} \myemph{eight augmentation techniques} on \myemph{four classifiers}, ranging from shallow architectures to BERT~\cite{devlin-etal-2019}, a popular pre-trained Transformer network.
We \added[id=T]{used} downsampled variants of the Kaggle Toxic Comment Classification Challenge dataset ~(\citealt{jigsaw2018toxic}; ~\S\ref{sec:methodology}) as our seed dataset. We 
\added[id=T]{focused} on the \cls{threat} class, but also replicated our results on another toxic class (\S\ref{sec:identity-hate}).
%
With some classifiers, we \added[id=T]{reached} the same F1-score as when training on the original dataset, which is 20x larger.
However, performance \added[id=T]{varied} markedly between classifiers.

We \added[id=T]{obtained} the highest overall results with BERT, increasing the F1-score up to 21\% compared to training on seed data alone. However, augmentation using a fine-tuned \gpt (\S\ref{sec:gpt2})~ -- a pre-trained Transformer language model~\cite{radford2019language} -- \added[id=T]{reached} almost BERT-level performance even with shallow classifiers. 
Combining multiple augmentation techniques, such as adding majority class sentences to minority class documents (\S\ref{sec:sentence-addition}) and replacing \emph{subwords} with embedding-space neighbors~\cite{heinzerling2018bpemb} (\S\ref{sec:bpemb}), \added[id=T]{improved} performance on all classifiers.
We discuss the interplay of performance and computational requirements like memory and run-time costs (\S\ref{sec:computational-requirements}).
We release our source code.\footnote{\url{https://github.com/ssg-research/language-data-augmentation}}
\section{Preliminaries}
\label{sec:background}

\myparagraph{Data augmentation} arises naturally from the problem of filling in missing values~\cite{tanner1987calculation}. In classification, data augmentation is applied to available training data.
Classifier performance is measured on a separate (non-augmented) test set~\cite{krizhevsky2012imagenet}.
Data augmentation can decrease \myemph{overfitting}~\cite{wong-etal-2016, shorten-khoshgoftaar-2019}, and broaden the input feature range by increasing the vocabulary \cite{fadaee-etal-2019}.





\myparagraph{Simple oversampling} is
the \added[id=T]{most basic} augmentation technique: copying minority class datapoints to appear multiple times. This increases the relevance of minority class features for computing the loss during training \cite{chawla-etal-2002}.

\myparagraph{EDA}
is a prior technique combining four text transformations to improve classification with CNN and RNN architectures~\cite{wei-zou-2019}.
It uses (i) synonym replacement from WordNet (\S\ref{sec:wordnet}), (ii) random insertion of a synonym, (iii) random swap of two words, and (iv) random word deletion.

\myparagraph{Word replacement} has been applied in several data augmentation studies \cite{Zhangetal2015, wang-yang-2015, xie-etal-2017, wei-zou-2019, fadaee-etal-2019}.
We \added[id=T]{compared} four techniques, two based on semantic knowledge bases (\S\ref{sec:wordnet}) and two on pre-trained (sub)word embeddings (\S\ref{sec:embedding-substitutions}).




\myparagraph{Pre-trained Transformer networks} feature prominently in state-of-the-art \added[id=T]{NLP research}.
\added[id=T]{They are able} to learn contextual embeddings, 
which depend on neighboring subwords~\cite{devlin-etal-2019}.
\emph{Fine-tuning} -- adapting \added[id=T]{the} weights of a pre-trained Transformer to a specific corpus -- has been highly effective in improving classification performance~\cite{devlin-etal-2019} and language modeling~\cite{radford2019language,aidungeon2,branwen2019gpt}. 
State-of-the-art networks are trained on large corpora: 
GPT-2's corpus contains 8M web pages, while 
BERT's training corpus contains 3.3B words. 


\section{Methodology}
\label{sec:methodology}

We now describe the data (\ref{sec:dataset}), augmentation techniques (\ref{sec:data-augmentation-techniques}), and classifiers (\ref{sec:classifiers}) we \added[id=T]{used}.

\subsection{Dataset}
\label{sec:dataset}

We used \myemph{Kaggle's toxic comment classification challenge} dataset~\cite{jigsaw2018toxic}.
It contains human-labeled English Wikipedia comments in six different classes of toxic language.\footnote{Although one class is specifically called \cls{toxic}, all six represent types of toxic language. See Appendix A.}
The median length of a document is three sentences, but the distribution is heavy-tailed (Table~\ref{tab:data-stats}). 

\begin{table}[tbh]
  \begin{center}
    \begin{tabular}{ccccccc}
    \hline
Mean & Std. & Min & Max & 25\% & 50\% & 75\%\\
4 & 6 & 1 & 683 & 2 & 3 & 5\\    
    \hline
    \end{tabular}
    \caption{Document lengths (number of sentences; tokenized with NLTK sent\_tokenize~\cite{NLTK09}).}
    \label{tab:data-stats}
  \end{center}
\end{table}

Some classes are severely under-represented:  
e.g., 478 examples of \cls{threat} vs. 159093 non-\cls{threat} examples.
Our experiments \added[id=M]{concern} binary classification, where one class is the \myemph{minority class} and all remaining documents belong to the \myemph{majority class}.
We \added[id=M]{focus} on \cls{threat} as the minority class, as it poses the most challenge for automated analysis in this dataset~\cite{van2018challenges}.
To confirm our results, we also \added[id=T]{applied} the best-performing techniques on a different type of toxic language, the \cls{identity-hate} class (\S\ref{sec:identity-hate}).

Our goal is to understand how data augmentation improves performance under extreme data scarcity in the minority class (\cls{threat}). 
To simulate this, we derive our seed dataset (\seed) from the full data set (\gold) via \myemph{stratified bootstrap sampling}~\citep{bickel1984asymptotic} to reduce the dataset size $k$-fold. 
We \added[id=T]{replaced} newlines, tabs and repeated spaces with single spaces, and lowercased each dataset.
We \added[id=T]{applied} data augmentation techniques on \seed with $k$-fold oversampling of the minority class,
and \added[id=T]{compared} each classifier architecture (\S\ref{sec:classifiers}) trained on \seed, \gold, and the augmented datasets.
We \added[id=T]{used} the original test dataset (\test) for evaluating performance.
We detail the dataset sizes in Table ~\ref{tab:data-stats-gold}. 



\begin{table}[tbh]
  \begin{center}
    \begin{tabular}{lccc}
    \hline
     & \textsc{Gold Std.} &  \seed & \test \\ 
    Minority & 478 & 25 & 211 \\ 
    Majority & 159,093 & 7955 & 63,767 \\ 
    \hline
    \end{tabular}
    \caption{Number of documents (minority: \cls{threat})}
    \label{tab:data-stats-gold}
  \end{center}
\end{table}


\noindent{\textbf{Ethical considerations}.}
We \added[id=T]{used only} public 
datasets,
and did not involve human subjects. 

\subsection{Data augmentation techniques}
\label{sec:data-augmentation-techniques}

We \added[id=T]{evaluated} six data augmentation techniques on four classifiers (Table~\ref{tab:pretrained-corpus}). We describe each augmentation technique (below) and classifier (\S\ref{sec:classifiers}). 
%
\begin{table*}[tbh]
\begin{center}
\begin{tabular}{lcccc}
\hline
\textbf{Augmentation} & \textbf{Type} & \textbf{Unit} &  \textbf{\#Parameters} & \textbf{Pre-training Corpus}  \\ 
\hline
\add & Non-toxic corpus & Sentence & NA & NA \\
\ppdb & Knowledge Base & N-gram & NA & NA \\ 
\wn & Knowledge Base & Word & NA & NA \\
\gensim & GloVe & Word & 30M & Twitter  \\
\bpemb & GloVe & Subword & 0.5M & Wikipedia  \\
\gpt & Transformer & Subword & 117M & WebText \\
\hline
\hline
\textbf{Classifier} & \textbf{Model Type} &  \textbf{Unit} & \textbf{\#Parameters} & \textbf{Pre-training Corpus} \\
\hline
\charlr & Logistic regression & Character & 30K & - \\
\wordlr & Logistic regression & Word & 30K & - \\
\cnn & Convolutional network & Word & 3M & - \\
\bert & Transformer & Subword & 110M & Wikipedia \& BookCorpus \\
\hline
\end{tabular}
\caption{Augmentation techniques and classifiers considered in this study.}
\label{tab:pretrained-corpus}
\end{center}
\end{table*}
%
%
%
%
%
%
%
For comparison, we also \added[id=T]{evaluated} simple oversampling (\simple) and EDA \cite{wei-zou-2019}, both reviewed in \S\ref{sec:background}.
Following the recommendation of \citet{wei-zou-2019} for applying EDA to small seed datasets, 
we \added[id=T]{used} $5\%$ augmentation probability, whereby each word has a $1 - 0.95^{4} \approx 19\%$ probability of being transformed by at least one of the four \eda~techniques. 

Four of \added[id=T]{the} six techniques are based on replacing words with semantically close counterparts; two using semantic knowledge bases (\S\ref{sec:wordnet}) and two pre-trained embeddings (\S\ref{sec:embedding-substitutions}).
We \added[id=T]{applied} 25\% of all possible replacements with these techniques, 
which is close to the recommended substitution rate in \eda. 
\added[id=M]{For short documents we \added[id=T]{ensured that} at least one substitution is always selected.}
We also \added[id=T]{added} majority class material to minority class documents (\S\ref{sec:sentence-addition}), and \added[id=T]{generated} text with the \gpt~ language model fine-tuned on \seed~ (\S\ref{sec:gpt2}).




\subsubsection{Substitutions from a knowledge base}
\label{sec:wordnet}

\noindent{\textbf{WordNet}}
is a semantic knowledge base containing various properties of \myemph{word senses}, which correspond to word meanings \cite{miller-1995}. We \added[id=T]{augmented} \seed by replacing words with random \myemph{synonyms}.
While EDA also uses WordNet synonyms (\S\ref{sec:background}), we additionally \added[id=T]{applied} word sense disambiguation \cite{Navigli2009} and inflection.


For word sense disambiguation we \added[id=T]{used} \myemph{simple Lesk} from PyWSD \cite{pywsd14}. As a variant of the Lesk algorithm \cite{Lesk1986} it relies on overlap in definitions and example sentences (both provided in WordNet), compared between each candidate sense and words in the context.

Word senses appear as uninflected \myemph{lemmas},
which we \added[id=T]{inflected} using a dictionary-based technique.
We lemmatized and annotated a large corpus with NLTK \cite{NLTK09}, and mapped each $<${lemma}, tag{}$>$ combination to its most common surface form.
The corpus contains $8.5$ million short sentences ($\leq 20$ words) from multiple open-source corpora (see Appendix E).
We designed it to have both a large vocabulary for wide coverage ($371125$ lemmas), and grammatically simple sentences to maximize correct tagging.

\noindent{\textbf{Paraphrase Database (\ppdb)}}
was collected from bilingual parallel corpora on the premise that English phrases translated identically to another language tend to be paraphrases \cite{Ganitkevitchetal2013, Pavlicketal2015}.
We \added[id=T]{used} phrase pairs tagged as \myemph{equivalent}, constituting $245691$ paraphrases altogether.
We \added[id=T]{controlled} substitution by grammatical context as specified in PPDB. In single words this is the part-of-speech tag; whereas in multi-word paraphrases it also contains the syntactic category that appears after the original phrase in the PPDB training corpus.
We \added[id=T]{obtained} grammatical information with the \myemph{Spacy}\footnote{\url{https://spacy.io/}} parser.

\subsubsection{Embedding neighbour substitutions}
\label{sec:embedding-substitutions}

Embeddings can be used to map units to others with a similar occurrence distribution in a training corpus \cite{mikolov-etal-2013}.
We \added[id=T]{considered} two alternative pre-trained embedding models.
For each model, we \added[id=T]{produced} top-\added[id=M]{$10$} nearest embedding neighbours \added[id=M]{(cosine similarity) }of each word selected for replacement, and randomly \added[id=T]{picked} the new word from these. 

\noindent{\textbf{Twitter word embeddings (\gensim)}}
\label{sec:glove}
\cite{pennington-etal-2014} were obtained from a Twitter corpus,\footnote{We use $25$-dimensional GloVe-embeddings from: \url{https://nlp.stanford.edu/projects/glove/}}
and we deployed these via Gensim~\cite{rehurek_lrec}.

\noindent{\textbf{Subword embeddings (\bpemb)}}
\label{sec:bpemb}
have emerged as a practical pre-processing tool
for overcoming the challenge of low-prevalence words ~\cite{sennrich2016neural}.
They have been applied in Transformer algorithms, including WordPiece~\cite{wu2016google} for \bert~\cite{devlin-etal-2019}, and BPE~\cite{sennrich2016neural} for GPT-2~\cite{radford2019language}.
\bpemb~\cite{heinzerling2018bpemb} provides pre-trained GloVe 
embeddings, constructed by applying SentencePiece~\cite{kudo2018sentencepiece}
on the English Wikipedia. 
We use $50$-dimensional \bpemb-embeddings with vocabulary size 10,000. 

\subsubsection{Majority \added[id=T]{class} sentence addition (\add)} 
\label{sec:sentence-addition}

Adding unrelated material to the training data can be beneficial by making relevant features stand out \cite{wong-etal-2016, shorten-khoshgoftaar-2019}. We \added[id=T]{added} a random sentence from a \emph{majority class} document in \seed to a random position in a copy of each minority class training document. 

\subsubsection{\gpt~conditional generation} 
\label{sec:gpt2}

GPT-2 is a Transformer language model pre-trained on a large collection 
of Web documents. 
We \added[id=T]{used} the 110M parameter GPT-2 model 
from the Transformers library~\cite{wolf2019Transformers}
We discuss parameters in Appendix~F.
We \added[id=T]{augmented} as follows ($N$-fold oversampling):
\begin{algorithmic}
\STATE{1.} $\hat{G} \leftarrow$ briefly \textbf{train} \gpt~on minority class documents in \seed. 
\STATE{2.}  \textbf{generate} $N-1$ novel documents $\hat{\mathbf{x}} \leftarrow \hat{G} (\mathbf{x})$ for all minority class samples $\mathbf{x}$ in \seed.
\STATE{3.}  \textbf{assign} the minority class label to all documents $\hat{\mathbf{x}}$
\STATE{4.}  \textbf{merge} $\hat{\mathbf{x}}$ with \seed. 
\end{algorithmic}

%

\subsection{Classifiers}
\label{sec:classifiers}


\myparagraphS{\charlr~and \wordlr}
We \added[id=T]{adapted} the logistic regression pipeline from the Wiki-detox project \cite{Wulczynetal2017}.\footnote{\url{https://github.com/ewulczyn/wiki-detox/blob/master/src/modeling/get_prod_models.py}}
We \added[id=T]{allowed} n-grams in the range 1--4, \added[id=T]{and kept} the default parameters: TF-IDF normalization, vocabulary size at $10,000$ and parameter $C = 10$ (inverse regularization strength). 

\myparagraphS{\cnn} We \added[id=T]{applied} a word-based CNN model with $10$ kernels of sizes $3$, $4$ and $5$. Vocabulary size \added[id=T]{was} $10,000$ 
and embedding dimensionality $300$. For training, we \added[id=T]{used} the dropout probability of $0.1$, and the Adam optimizer \cite{kingma-ba-2014} with the learning rate of $0.001$.

\myparagraphS{\bert}
We used the pre-trained Uncased BERT-Base and trained the model with the training script from Fast-Bert.\footnote{\url{https://github.com/kaushaltrivedi/fast-bert/blob/master/sample_notebooks/new-toxic-multilabel.ipynb}}
We set maximum sequence length to 128 and mixed precision optimization level to O1.

\section{Results}
\label{sec:results}


We \added[id=T]{compared} precision and recall for the minority class (\cls{threat}), and the macro-averaged F1-score for each classifier and augmentation technique. (For brevity, we use ``F1-score" from now on.) The majority class F1-score \added[id=T]{remained} $1.00$ (two digit rounding) across all our experiments. 
All classifiers are binary, and we \added[id=T]{assigned} predictions to the class with the highest conditional probability.
We relax this assumption in \S\ref{sec:tradeoff}, to report  area under
the curve (AUC) values~\cite{murphy2012machine}. 

To validate our results, we \added[id=T]{performed} repeated experiments with the \myemph{common random numbers} technique~\cite{glasserman1992some}, by which we \added[id=T]{controlled} the sampling of \seed, 
initial random weights of classifiers, and the optimization procedure. 
We \added[id=T]{repeated} the experiments~\numreps~times, and report confidence intervals. 

%

\subsection{Results without augmentation}
\label{sec:results-gold}

We first show classifier performance on \gold~and \seed~in Table~\ref{tab:results-gold}. 
\citeauthor{van2018challenges}~(\citeyear{van2018challenges}) reported F1-scores for logistic regression and CNN classifiers on \gold. Our results are comparable. We also evaluate \bert, which is noticeably better on \gold, particularly in terms of \cls{threat} recall. 

\begin{table}[t]
  \begin{center}
    \begin{tabular}{lcccc}
    \hline    
    & \multicolumn{4}{c}{\gold} \\
    {} & {\charlr} &  {\wordlr} &  {\cnn} & {\bert}\\
    Precision & 0.61 & 0.43 & 0.60 & 0.54\\
    Recall & 0.34 & 0.36 & 0.33 & 0.54 \\
    F1 & 0.72 & 0.69 & 0.71 & 0.77\\
    \hline
    & \multicolumn{4}{c}{\seed} \\ 
    {} & {\charlr} &  {\wordlr} &  {\cnn} & {\bert}\\
    Precision & 0.64 & 0.47 & 0.41 & 0.00 \\
    Recall & 0.03 & 0.04 & 0.09 & 0.00 \\
    F1 & 0.52 & 0.53 & 0.57 & 0.50 \\
    \hline
    \end{tabular}
    \caption{Classifier performance on \gold~and \seed. 
    Precision and recall for \cls{threat}; F1-score macro-averaged from both classes.}
    \label{tab:results-gold}
  \end{center}
\end{table}

All classifiers \added[id=T]{had} significantly reduced F1-scores on \seed,  due to \added[id=T]{major} drops in \cls{threat} recall. In particular, \bert \added[id=T]{was} degenerate, assigning all documents to the majority class in all \numreps~repetitions.
\citeauthor{devlin-etal-2019} (\citeyear{devlin-etal-2019}) report that such behavior may occur on small datasets, 
but random restarts may help. 
In our case, random restarts did not impact \bert~performance on \seed. 

\subsection{Augmentations}
\label{sec:augmentations}

We applied all eight augmentation techniques (\S\ref{sec:data-augmentation-techniques}) to the minority class of \seed (\cls{threat}).
Each technique retains one copy of each \seed~document, and adds 19 synthetically generated documents per \seed~document. 
Table~\ref{tab:data-stats-augmented} summarizes augmented dataset sizes. 
We present our main results in Table~\ref{tab:results-augmented}. We first discuss classifier-specific observations, and then make general observations on each augmentation technique.

\begin{table}[h]
  \begin{center}
    \begin{tabular}{lcc}
    \hline
     & \seed & Augmented   \\ 
    Minority & 25 & 25$\rightarrow$500 \\ 
    Majority & 7955 & 7955 \\ 
    \hline
    \end{tabular}
    \caption{Number of documents in augmented datasets. We \added[id=T]{retained} original \seed~documents and \added[id=T]{expanded} the dataset with additional synthetic documents \added[id=T]{(minority: \cls{threat})}}
    \label{tab:data-stats-augmented}
  \end{center}
\end{table}

We \added[id=T]{compared} the impact of augmentations on each classifier, and therefore our performance comparisons below are local to each column (i.e., classifier). We identify the best performing technique for the three metrics and report the p-value when its effect is significantly better than the other techniques (based on one-sided paired t-tests, $\alpha=5\%$).\footnote{The statistical significance results apply to \myemph{this} dataset, but are indicative of the behavior of the techniques in general.}

\begin{table*}[!h]
  \begin{center}
    \begin{tabular}{llcccc}
    \hline
     \textbf{Augmentation} & \textbf{Metric} & \textbf{\charlr} & \textbf{\wordlr} & \textbf{\cnn} & \textbf{\bert}\\
     \hline
    \multirow{3}{*}{\shortstack[l]{\seed \\ \seedL}}&Precision & $\ta{0.68} \pm 0.22$ & $\ta{0.43} \pm 0.27$ & $\ta{0.45} \pm 0.14$ & $0.00 \pm 0.00$ \\
    &Recall & $0.03 \pm 0.02$ & $0.04 \pm 0.02$ & $0.08 \pm 0.05$ & $0.00 \pm 0.00$ \\
    &F1 (macro) & $0.53 \pm 0.02$ & $0.54 \pm 0.02$ & $0.56 \pm 0.03$ & $0.50 \pm 0.00$ \\
    \hline
    \multirow{3}{*}{\shortstack[l]{\simple \\ \simpleL}}&Precision & $\marg{0.67} \pm 0.07$ & $\marg{0.38} \pm 0.24$ & $0.40 \pm 0.08$ & $\ta{0.49} \pm 0.07$ \\
    &Recall & $0.16 \pm 0.03$ & $0.03 \pm 0.02$ & $0.07 \pm 0.03$ & $0.36 \pm 0.09$ \\
    &F1 (macro) & $0.63 \pm 0.02$ & $0.53 \pm 0.02$ & $0.56 \pm 0.02$ & $\marg{0.70} \pm 0.03$ \\
    \hline
    \multirow{3}{*}{\shortstack[l]{\eda \\ \citeauthor{wei-zou-2019}~(\citeyear{wei-zou-2019})}}&Precision & $\marg{0.66} \pm 0.06$ & $0.36 \pm 0.19$ & $0.26 \pm 0.09$ & $0.21 \pm 0.03$ \\
    &Recall & $0.13 \pm 0.03$ & $0.08 \pm 0.04$ & $0.07 \pm 0.01$ & $0.06 \pm 0.01$ \\
    &F1 (macro) & $0.61 \pm 0.02$ & $0.56 \pm 0.03$ & $0.55 \pm 0.01$ & $0.54 \pm 0.01$ \\
    \hline
    \multirow{3}{*}{\shortstack[l]{\add \\ \addL}}&Precision & $0.58 \pm 0.07$ & $0.36 \pm 0.21$ & $\marg{0.45} \pm 0.07$ & $0.36 \pm 0.04$ \\
    &Recall & $0.24 \pm 0.04$ & $0.06 \pm 0.04$ & $0.19 \pm 0.07$ & $0.52 \pm 0.07$ \\
    &F1 (macro) & $0.67 \pm 0.03$ & $0.55 \pm 0.03$ & $0.63 \pm 0.04$ & $\ta{0.71} \pm 0.01$ \\
    \hline
    \multirow{3}{*}{\shortstack[l]{\ppdb \\ \ppdbL}}&Precision & $0.16 \pm 0.08$ & $\marg{0.41} \pm 0.27$ & $0.37 \pm 0.09$ & $\marg{0.48} \pm 0.06$ \\
    &Recall & $0.10 \pm 0.03$ & $0.04 \pm 0.02$ & $0.08 \pm 0.04$ & $0.34 \pm 0.08$ \\
    &F1 (macro) & $0.56 \pm 0.02$ & $0.53 \pm 0.02$ & $0.57 \pm 0.02$ & $0.70 \pm 0.03$ \\
    \hline
    \multirow{3}{*}{\shortstack[l]{\wn \\ \wnL}}&Precision & $0.16 \pm 0.06$ & $0.36 \pm 0.24$ & $\marg{0.41} \pm 0.08$ & $0.47 \pm 0.08$ \\
    &Recall & $0.11 \pm 0.03$ & $0.05 \pm 0.03$ & $0.11 \pm 0.05$ & $0.29 \pm 0.07$ \\
    &F1 (macro) & $0.56 \pm 0.02$ & $0.54 \pm 0.02$ & $0.58 \pm 0.03$ & $0.68 \pm 0.03$ \\
    \hline
    \multirow{3}{*}{\shortstack[l]{\gensim \\ \gensimL}}&Precision & $0.15 \pm 0.04$ & $\marg{0.39} \pm 0.12$ & $0.38 \pm 0.08$ & $0.43 \pm 0.11$ \\
    &Recall & $0.14 \pm 0.03$ & $0.16 \pm 0.05$ & $0.18 \pm 0.06$ & $0.18 \pm 0.06$ \\
    &F1 (macro) & $0.57 \pm 0.02$ & $0.61 \pm 0.03$ & $0.62 \pm 0.03$ & $0.62 \pm 0.03$ \\
    \hline
    \multirow{3}{*}{\shortstack[l]{\bpemb \\ \bpembL}}&Precision & $0.56 \pm 0.07$ & $0.33 \pm 0.07$ & $0.25 \pm 0.07$ & $0.38 \pm 0.12$ \\
    &Recall & $0.22 \pm 0.03$ & $0.22 \pm 0.04$ & $0.37 \pm 0.08$ & $0.16 \pm 0.04$ \\
    &F1 (macro) & $0.66 \pm 0.02$ & $0.63 \pm 0.02$ & $0.64 \pm 0.03$ & $0.61 \pm 0.03$ \\
    \hline
    \multirow{3}{*}{\shortstack[l]{\gpt \\ \gptL}}&Precision & $0.45 \pm 0.08$ & $0.35 \pm 0.07$ & $0.31 \pm 0.08$ & $0.15 \pm 0.05$ \\
    &Recall & $\taa{0.33} \pm 0.04$ & $\taa{0.42} \pm 0.05$ & $\taa{0.46} \pm 0.10$ & $\taa{0.62} \pm 0.09$ \\
    &F1 (macro) & $\taa{0.69} \pm 0.02$ & $\taa{0.69} \pm 0.02$ & $\taa{0.68} \pm 0.02$ & $0.62 \pm 0.03$ \\
    \hline
    \end{tabular}
    \caption{Comparison of augmentation techniques for 20x augmentation on \seed/\cls{threat}´: 
    means for precision, recall and macro-averaged F1-score shown with standard deviations (\numreps~paired repetitions). 
    Precision and recall for \cls{threat}; F1-score macro-averaged from both classes.
    \marg{Bold} figures represent techniques that are either best, or \emph{not} significantly different ($\alpha=5\%$) from this best technique.
    \taa{Double underlines} indicate the best technique (for a given metric and classifier) significantly better ($\alpha=1\%$) than all other techniques.
    }
    \label{tab:results-augmented}
  \end{center}
\end{table*}    

\myparagraphS{\bert}
\simple~and \add~\added[id=T]{were} successful on \bert, raising the F1-score up to 21 percentage points above \seed to 0.71. But their impacts on \bert \added[id=T]{were} different: \add \added[id=T]{led} to increased recall, while \simple \added[id=T]{resulted} in increased precision. 
\ppdb precision and recall \added[id=T]{were} statistically indistinguishable from \simple, which indicates that it \added[id=T]{did} few alterations.  
\gpt \added[id=T]{led} to significantly better recall ($p < 10^{-5}$ for all pairings), even surpassing \gold. 
Word substitution methods like \eda, \wn, \gensim, and \bpemb \added[id=T]{improved} on \seed, but \added[id=T]{were} less effective than \simple \added[id=T]{in} both precision and recall. 
\citet{park2019thisiscompetition} found that \bert may perform poorly on out-of-domain samples. \bert~is reportedly unstable on adversarially chosen subword substitutions~\cite{sun2020adv}.
We suggest that non-contextual word embedding schemes may be sub-optimal for \bert
since its pre-training is not conducted with similarly noisy documents. 
We verified that reducing the number of replaced words was indeed beneficial for \bert~(Appendix~G).

\myparagraphS{\charlr}
\bpemb~and \add~\added[id=T]{were} effective at increasing recall, and reached similar increases in F1-score. \gpt~\added[id=T]{raised} recall to \gold~level ($p < 10^{-5}$ for all pairings), but precision \added[id=T]{remained} 16 percentage points below \gold. It \added[id=T]{led} to the best increase in F1-score: 16 percentage points above \seed~($p < 10^{-3}$ for all pairings).

\myparagraphS{\wordlr}
Embedding-based \bpemb and \gensim \added[id=T]{increased} recall by at least 13 percentage points, 
but the conceptually similar \ppdb~and \wn~\added[id=T]{were} largely unsuccessful. We suggest this discrepancy may be due to \wn~and \ppdb~relying on \emph{written standard} English, whereas toxic language tends to be more colloquial. \gpt~\added[id=T]{increased} recall and F1-score the most: 15 percentage points above \seed~($p < 10^{-10}$ for all pairings).

\myparagraphS{\cnn} \gensim~and \add~\added[id=T]{increased} recall by at least 10 percentage points. 
\bpemb \added[id=T]{led} to a large increase in recall, but with a drop in precision, 
possibly due to its larger capacity to make changes in text -- \gensim can only 
replace entire words that exist in the pre-training corpus. 
\gpt \added[id=T]{yielded} the largest increases in recall and F1-score~($p < 10^{-4}$ for all pairings).

We now discuss each augmentation technique.

\myparagraph{\simple} 
emphasizes the features of original minority documents in \seed, which generally \added[id=T]{resulted} in fairly high precision. \added[id=M]{On \wordlr, \simple is analogous to increasing the weight of words that appear in minority documents.}

\myparagraph{\eda} 
\added[id=T]{behaved} similarly to \simple on \charlr, \wordlr~and \cnn; but markedly worse on \bert. 

\myparagraph{\add} 
reduces the classifier's sensitivity to irrelevant material by adding majority class sentences to minority class documents. 
On \wordlr, \add is analogous to reducing the weights of majority class words.
\add \added[id=T]{led} to a marginally better F1-score than any other technique on \bert.

\myparagraph{Word replacement} \added[id=T]{was} more effective with \gensim~and \bpemb than with \ppdb or \wn. 
\ppdb and \wn generally replace few words \added[id=T]{per document}, which often \added[id=T]{resulted} in similar performance to \simple. 
\bpemb \added[id=T]{was} generally the most effective among these techniques. 


\myparagraph{\gpt} \added[id=T]{had} the best improvement overall, leading to significant increases in recall across all classifiers,
and the highest F1-score on all but \bert.
The increase in recall can be attributed to \gpt's capacity for introducing \emph{novel phrases}.
We corroborated this hypothesis by measuring the overlap between the original and augmented test sets and an offensive/profane word list from von Ahn.\footnote{\url{https://www.cs.cmu.edu/~biglou/resources/}} \gpt augmentations increased the intersection cardinality by $260\%$ from the original; compared to only $84\%$ and $70\%$ with the next-best performing augmentation techniques (\add and \bpemb, respectively). This demonstrates that \gpt significantly increased the vocabulary range of the training set, specifically with offensive words likely to be relevant for toxic language classification.
However, there is a risk that human annotators might not label \gpt-generated documents as toxic.
Such \myemph{label noise} may decrease precision.
(See Appendix H, Table 22 for example augmentations that display the behavior of \gpt and other techniques.)


\subsection{Mixed augmentations}

In \S\ref{sec:augmentations} we saw that the effect of augmentations differ across classifiers. A natural question is whether it is beneficial to combine augmentation techniques.
For all classifiers except \bert, 
the best performing techniques 
\added[id=T]{were} \gpt, \add, and \bpemb (Table \ref{tab:results-augmented}).
They also represent each of our augmentation types (\S\ref{sec:data-augmentation-techniques}), \bpemb~having the highest performance among the four word replacement techniques (\S\ref{sec:wordnet}--\S\ref{sec:embedding-substitutions}) in these classifiers.

We \added[id=T]{combined} the techniques by merging augmented documents in equal proportions.
In \abg, we \added[id=T]{included} documents generated by \add, \bpemb~or \gpt. 
Since \add and \bpemb impose significantly lower computational and memory requirements than \gpt, and require no access to a GPU (Appendix~C),
we also \added[id=T]{evaluated} combining \added[id=T]{only} \add and \bpemb (\ab).

\abg \added[id=T]{outperformed} all other techniques (in F1-score) on \charlr~and \cnn~with statistical significance, while being marginally better on \wordlr. 
On \bert, \abg \added[id=T]{achieved} a better F1-score and precision than \gpt alone ($p < 10^{-10}$), \added[id=M]{and a better recall ($p < 0.05$)}.
\added[id=M]{\abg }
\added[id=T]{was} better than \ab in recall on \wordlr and \cnn, 
while the precision \added[id=T]{was} comparable. 

\begin{table}[!b]
  \begin{center}
    \begin{tabular}{lcccc}
    \hline
    & \multicolumn{4}{c}{\ab} \\
      & \charlr & \wordlr & \cnn & \bert \\
    Precision & $0.56$ & $0.37$ & $0.33$ & $0.41$ \\
    Recall & $0.26$ & $0.18$ & $0.36$ & $0.36$ \\
    F1 & $0.68$ & $0.62$ & $0.67$ & $0.69$ \\
    \hline
    & \multicolumn{4}{c}{\abg} \\
      & \charlr & \wordlr & \cnn & \bert \\
    Precision & $0.48$ & $\marg{0.37}$ & $0.31$ & $0.28$ \\
    Recall & $\underline{\underline{\textbf{0.36}}}$ & $0.39$ & $\underline{\underline{\textbf{0.52}}}$ & $\ta{0.65}$ \\
    F1 & $\underline{\underline{\textbf{0.70}}}$ & $\marg{0.69}$ & $\taa{0.69 }$ & $0.69$ \\
    \hline
    \end{tabular}
    \caption{Effects of mixed augmentation (20x) on \seed/\cls{threat}  
    (Annotations as in Table~\ref{tab:results-augmented}).
    Precision and recall for \cls{threat}; F1-score macro-averaged from both classes.}
    \label{tab:results-combinations}
  \end{center}
\end{table}

Augmenting with \abg \added[id=T]{resulted} in similar performance as \gold on \wordlr, \charlr~and \cnn~(Table~\ref{tab:results-gold}). 
Comparing~Tables~\ref{tab:results-augmented} and \ref{tab:results-combinations}, 
it is clear that much of the performance improvement \added[id=T]{came} from
the increased vocabulary coverage of \gpt-generated documents. 
Our results suggest that in certain types of data like toxic language, 
consistent labeling may be more important than wide coverage in dataset collection, 
since automated data augmentation can increase the coverage of language.
Furthermore, \charlr~trained with \abg \added[id=T]{was} comparable (no statistically significant difference) to the best results obtained with \bert (trained with \add, $p > 0.2$ on all metrics). 

\subsection{Average classification performance}
\label{sec:tradeoff}

The results in Tables \ref{tab:results-augmented} and \ref{tab:results-combinations} focus on precision, recall and the F1-score of different models and augmentation techniques where the probability threshold for determining the positive or negative class is $0.5$. In general the level of precision and recall are adapted based on the use case for the classifier. Another \replaced[id=M]{}{more} general evaluation of a classifier is based on the \rocauc metric, which is the area under the curve for a plot of true-positive rate versus the false-positive rate for a range of thresholds varying  over $\lbrack 0, 1\rbrack$. Table \ref{tab:roc_auc} shows the \rocauc scores for each of the classifiers for the best augmentation techniques from Tables~\ref{tab:results-augmented} and~\ref{tab:results-combinations}.

\bert with \abg \added[id=T]{gave} the best \rocauc value of $0.977$ which is significantly higher than \bert with any other augmentation technique ($p < 10^{-6}$). \cnn 
\added[id=T]{exhibited} a similar pattern: \abg \added[id=T]{resulted} in the best \rocauc compared to the other augmentation techniques ($p< 10^{-6}$). 
\added[id=M]{For \wordlr, \rocauc was highest for \abg, but the difference to \gpt was not statistically significant ($p > 0.05$).}
In the case of \charlr, none of the augmentation techniques \added[id=T]{improved} on \seed ($p < 0.05$). \charlr \added[id=T]{produced} a more consistent \replaced[id=M]{averaged}{generalisation} performance across all augmentation methods with \rocauc values varying between $(0.958, 0.973)$, compared to variations across all augmentation techniques of $(0.792, 0.962)$ and $(0.816, 0.977)$ for \cnn and \bert respectively. 

\par
\begin{table}[!h]
  \begin{center}
\resizebox{\linewidth}{!}{
\begin{tabular}{lcccc}
\hline
                                 & \charlr   & \wordlr    & \cnn        & \bert       \\
\hline
\seed     & \ta{0.973} & 0.968      & 0.922       & 0.816 \\
\simple   & 0.972       & 0.937      & 0.792  & 0.898 \\
\add      & 0.958       & 0.955      & 0.904       & 0.956 \\
\bpemb    & 0.968       & 0.968      & 0.940       & 0.868 \\
\gpt     & 0.969       & \ta{0.973} & 0.953       & 0.964 \\
\abg   & 0.972       & \ta{0.973} & \taa{0.962} & \taa{0.977} \\
\hline

\end{tabular}}
    \caption{Comparison of \rocauc for augmentation (20x) on \seed/\cls{threat} (Annotations as in Table~\ref{tab:results-augmented}).}
     \label{tab:roc_auc}
     \end{center}
    
\end{table}

Our results highlight a difference between the results in Tables~\ref{tab:results-augmented} and~\ref{tab:results-combinations}: while \simple reached a high F1-score on \bert, our results on \rocauc~highlight that such performance may not hold while varying the decision threshold.
We observe that a combined augmentation method such as \abg provides an increased ability to vary \added[id=T]{the} decision threshold for the more complex classifiers such as \cnn and \bert. 
Simpler models \added[id=T]{performed} consistently across different augmentation techniques.



\subsection{Computational requirements}
\label{sec:computational-requirements}

\bert 
has significant computational requirements (Table~\ref{tab:clf-memory}). 
Deploying \bert~on common EC2 instances requires 13 GB GPU memory. 
\abg on EC2 requires 4 GB GPU memory for approximately 100s (for 20x augmentation). 
All other techniques take only a few seconds on ordinary desktop computers (See Appendices C--D
for additional data on computational requirements).

\begin{table}[h]
  \begin{center}
    \begin{tabular}{lcccc}
    \hline
     & \add & \bpemb & \gpt & \abg\\    
    \hline
    CPU & - & 100 & 3,600 & 3,600 \\
    GPU & - & - & 3,600 & 3,600 \\
    \hline
    & \charlr & \wordlr & \cnn & \bert \\    
    \hline
    CPU & 100 & 100 & 400 & 13,000 \\
    GPU & 100 & 100 & 400 & 13,000 \\
    \hline
    \end{tabular}
    \caption{
    Memory (MB) required for augmentation techniques and classifiers. Rounded to nearest 100 MB.
}
    \label{tab:clf-memory}
  \end{center}
\end{table}


\subsection{Alternative toxic class}
\label{sec:identity-hate}
In order to see whether our results described so far generalize beyond \cls{threat}, we repeated our experiments using another toxic language class, \cls{identity-hate}, as the minority class. Our results for \cls{identity-hate} are in line with those for \cls{threat}. All classifiers \added[id=T]{performed} poorly on \seed due to very low recall. Augmentation with simple techniques \added[id=T]{helped} \bert gain more than 20 percentage points for the F1-score. Shallow classifiers \added[id=T]{approached} \bert-like performance with appropriate augmentation.
We present further details in Appendix B.

\section{Related work}
\label{sec:related}

Toxic language classification has been conducted in a number of studies \cite{Schmidt:Wiegand2017, Davidsonetal2017, Wulczynetal2017, Grondahletal2018, qian2019benchmark, breitfeller2019finding}.
NLP applications of data augmentation include text classification \cite{ratner-etal-2017, wei-zou-2019, mesbah2019training}, user behavior categorization \cite{wang-yang-2015}, dependency parsing \cite{vania-etal-2019}, and machine translation \cite{fadaee-etal-2019, xia-etal-2019-generalized}.
Related techniques are also used in automatic paraphrasing \cite{Madnani:Dorr2010, Lietal2018} and writing style transfer \cite{Shenetal2017, Shettyetal2018, Mahmoodetal2019}.

\citet{Huetal2017} \added[id=T]{produced} text with controlled target attributes via variational autoencoders.
\citeauthor{mesbah2019training}~(\citeyear{mesbah2019training}) \added[id=T]{generated} artificial sentences for adverse drug reactions 
using
Reddit and Twitter data. 
Similarly to their work, we \added[id=T]{generated} novel toxic sentences from a language model. 
\citeauthor{petroni2019language}~(\citeyear{petroni2019language}) compared several pre-trained language models on their ability to understand factual and commonsense reasoning. BERT models consistently outperformed other language models. \citeauthor{petroni2019language} suggest that large pre-trained language models may become alternatives to knowledge bases in \added[id=T]{the} future. 


\section{Discussion and conclusions}
\label{sec:discussion}

Our results highlight the relationship between classification performance and computational overhead. 
Overall, \bert~\added[id=T]{performed the} best with \added[id=T]{data} augmentation. 
However, it is highly resource-intensive (\S\ref{sec:computational-requirements}).
\abg \added[id=T]{yielded} almost \bert-level F1- and \rocauc scores on \myemph{all} classifiers.
While using \gpt is more expensive than other augmentation techniques, it has significantly less requirements than \bert.
Additionally, augmentation is a \myemph{one-time upfront cost} in contrast to ongoing costs for classifiers.
Thus, the trade-off between performance and computational resources can 
influence which technique is optimal in a given setting.

We identify the following further topics that we leave for future work.


\myparagraphS{\seed~coverage}
Our results show that data augmentation can increase coverage, 
leading to better toxic language classifiers when starting with \myemph{very small} seed datasets.
The effects of data augmentation will likely differ with larger seed datasets.

\myparagraphS{Languages} Some augmentation techniques are limited in their applicability across languages. \gpt,~\wn, \ppdb~and \gensim~are available for certain other languages, but with less coverage than in English. \bpemb~is nominally available in 275 languages, but has not been thoroughly tested on less prominent languages.

%

\myparagraphS{Transformers} 
BERT has inspired work on other pre-trained Transformer classifiers, leading to 
better classification performance~\cite{liu2019roberta,lewis2019bart} and better trade-offs between memory consumption and classification performance~\cite{sanh2019distilbert,jiao2019tinybert}. 
Exploring the effects of augmentation on these Transformer classifiers is left for future work.


\myparagraphS{Attacks}
Training classifiers with augmented data may \added[id=T]{influence} their vulnerability for model extraction attacks~\cite{tramer2016stealing,krishna2020thieves}, 
model evasion~\cite{Grondahletal2018}, or backdoors~\cite{schuster2020humpty}. 
We leave such considerations for future work. 




\section*{Acknowledgments}
We thank Jonathan Paul Fernandez Strahl, Mark van Heeswijk, and Kuan Eeik Tan for valuable discussions related to the project, and Karthik Ramesh for his help with early experiments. We also thank Prof. Yaoliang Yu for providing compute resources for early experiments.
Tommi Gr\"{o}ndahl was funded by the Helsinki Doctoral Education Network in Information and Communications Technology (HICT).

\bibliographystyle{acl_natbib}
\bibliography{anthology,emnlp2020}

\appendix

\cleardoublepage

\appendix

\section{Class overlap and interpretation of ``toxicity''}
\label{app:data}

Kaggle's \myemph{toxic comment classification challenge} dataset\footnote{\url{https://www.kaggle.com/c/jigsaw-toxic-comment-classification-challenge}} contains six classes, one of which is called \cls{toxic}. But all six classes represent examples of toxic speech: \cls{toxic}, \cls{severe toxic}, \cls{obscene}, \cls{threat}, \cls{insult}, and \cls{identity-hate}. Of the \cls{threat} documents in the full training  dataset (\gold), $449/478$ overlap with \cls{toxic}. For \cls{identity-hate}, overlap with \cls{toxic} is $1302/1405$. Therefore, in this paper, we use the term \myemph{toxic} more generally, subsuming \cls{threat} and \cls{identity-hate} as particular types of toxic speech. To confirm that this was a reasonable choice, we manually examined the $29$ \cls{threat} datapoints not overlapping with \cls{toxic}. All of these represent genuine threats, and are hence toxic in the general sense.

\section{The ``Identity hate" class}
\label{app:other}

\begin{table}[tbh]
  \begin{center}
    \begin{tabular}{lccc}
    \hline
     & \textsc{Gold Std.} &  \seed &  \test \\
    Minority &  1,405 & 75 & 712 \\
    Majority & 158,166 & 7,910 & 63,266 \\
    \hline
    \end{tabular}
    \caption{Corpus size for \cls{identity-hate} (minority) and \cls{non-identity-hate} (majority).}
    \label{tab:data-stats-classes-identity}
  \end{center}
\end{table}

\begin{table}[tbh]
  \begin{center}
    \begin{tabular}{lcccc}
    \hline    
    & \multicolumn{4}{c}{\gold} \\
    {} & {Char} &  {Word} &  {CNN} & {BERT}\\
    Precision  & 0.64 & 0.54 & 0.70 & 0.55 \\
    Recall     & 0.40 & 0.31 & 0.20 & 0.62 \\
    F1 (macro) & 0.74 & 0.69 & 0.65 & 0.79 \\
    \hline
    \end{tabular}
    \caption{Classifier performance on \gold. 
    Precision and recall for \cls{identity-hate}; F1-score macro-averaged from both classes.} 
    \label{tab:results-gold-identity}
  \end{center}
\end{table}

To see if our results generalize beyond \cls{threat}, we experimented on the \cls{identity-hate} class in Kaggle's toxic comment classification dataset. 
Again, we used a 5\% stratified sample of \gold as \seed. 
We first show the number of samples in \gold, \seed and \test in Table~\ref{tab:data-stats-classes-identity}.  
There are approximately 3 times more minority-class samples in \cls{identity-hate} than in \cls{threat}. 
Next, we show classifier performance on \gold/\cls{identity-hate} in Table~\ref{tab:results-gold-identity}. 
The results closely resemble those on \gold/\cls{threat} in  Table~4 (\S4.1).

We compared \seed~and \simple~with the techniques that had the highest performance on \cls{threat}: \add, \bpemb, \gpt, and their combination \abg. 
Table~\ref{tab:results-identity} shows the results.

Like in \cls{threat}, \bert~performed the poorest on \seed, with the lowest recall ($0.06$).
All techniques decreased precision from \seed, and all increased recall except \simple~with \cnn. With \simple, the F1-score increased with \charlr~($0.12$) and \bert~($0.21$), but not \wordlr~($0.01$) or \cnn~($-0.04$). This is in line with corresponding results from \cls{threat} (\S4.2, Table 6): \simple~did not help either of the word-based classifiers (\wordlr, \cnn) but helped the character- and subword-based classifiers (\charlr, \bert).

Of the individual augmentation techniques, \add~increased the F1-score the most with \charlr~($0.15$) and \bert~($0.20$); and \gpt~increased it the most with \wordlr~($0.07$) and \cnn~($0.07$).
Here again we see the similarity between the two word-based classifiers, and the two that take inputs below the word-level.
Like in \cls{threat}, \simple~and \add~achieved close F1-scores with \bert, but with different relations between precision and recall.
\bpemb~was not the best technique with any classifier, but increased F1-score everywhere except in \cnn, where precision dropped drastically.

In the combined \abg~technique, \wordlr~and~\cnn reached their highest F1-score increases ($0.08$ and $0.07$, respectively).
With \charlr F1-score was also among the highest, but did not reach \add.
Like with \cls{threat}, \abg~increased precision and recall more than \gpt alone.

Overall, our results on \cls{identity-hate} closely resemble those we received in \cls{threat}, resulting in more than 20 percentage point increases in the F1-score for \bert on augmentations with \simple and \add.
Like in \cls{threat}, the impact of most augmentations was greater on \charlr~than on \wordlr or \cnn.
Despite their similar F1-scores in \seed, \charlr exhibited much higher precision, which decreased but remained generally higher than with other classifiers. Combined with an increase in recall to similar or higher levels than with other classifiers, \charlr~reached \bert-level performance with proper data augmentation.

\begin{table*}[tbh]
  \begin{center}
    \begin{tabular}{llcccc}
    \hline
     \textbf{Augmentation} & \textbf{Metric} & \textbf{\charlr} & \textbf{\wordlr} & \textbf{\cnn} & \textbf{\bert}\\
    \hline    
    \multirow{3}{*}{\shortstack[l]{\seed \\ \seedL}}&Precision & $0.85 \pm 0.04$ & $0.59 \pm 0.05$ & $0.52 \pm 0.08$ & $0.65 \pm 0.46$ \\
    &Recall & $0.11 \pm 0.04$ & $0.12 \pm 0.03$ & $0.11 \pm 0.04$ & $0.06 \pm 0.10$ \\
    &F1 (macro) & $0.60 \pm 0.03$ & $0.60 \pm 0.02$ & $0.59 \pm 0.02$ & $0.54 \pm 0.08$ \\
    \hline
    \multirow{3}{*}{\shortstack[l]{\simple \\ \simpleL}}&Precision & $0.61 \pm 0.02$ & $0.54 \pm 0.04$ & $0.27 \pm 0.06$ & $0.52 \pm 0.06$ \\
    &Recall & $0.34 \pm 0.04$ & $0.14 \pm 0.03$ & $0.07 \pm 0.01$ & $0.50 \pm 0.06$ \\
    &F1 (macro) & $0.72 \pm 0.02$ & $0.61 \pm 0.02$ & $0.55 \pm 0.01$ & $0.75 \pm 0.01$ \\
    \hline
    \multirow{3}{*}{\shortstack[l]{\add \\ \addL}}&Precision & $0.54 \pm 0.04$ & $0.54 \pm 0.05$ & $0.43 \pm 0.05$ & $0.43 \pm 0.05$ \\
    &Recall & $0.47 \pm 0.05$ & $0.21 \pm 0.03$ & $0.21 \pm 0.04$ & $0.58 \pm 0.08$ \\
    &F1 (macro) & $0.75 \pm 0.01$ & $0.65 \pm 0.01$ & $0.64 \pm 0.02$ & $0.74 \pm 0.01$ \\
    \hline
    \multirow{3}{*}{\shortstack[l]{\bpemb \\ \bpembL}}&Precision & $0.43 \pm 0.04$ & $0.30 \pm 0.03$ & $0.15 \pm 0.05$ & $0.29 \pm 0.06$ \\
    &Recall & $0.38 \pm 0.04$ & $0.29 \pm 0.01$ & $0.32 \pm 0.05$ & $0.23 \pm 0.03$ \\
    &F1 (macro) & $0.70 \pm 0.01$ & $0.64 \pm 0.01$ & $0.59 \pm 0.02$ & $0.62 \pm 0.02$ \\
    \hline
    \multirow{3}{*}{\shortstack[l]{\gpt \\ \gptL}}&Precision & $0.41 \pm 0.05$ & $0.30 \pm 0.03$ & $0.33 \pm 0.08$ & $0.22 \pm 0.05$ \\
    &Recall & $0.34 \pm 0.04$ & $0.39 \pm 0.03$ & $0.34 \pm 0.09$ & $0.59 \pm 0.06$ \\
    &F1 (macro) & $0.68 \pm 0.01$ & $0.67 \pm 0.01$ & $0.66 \pm 0.01$ & $0.65 \pm 0.02$ \\
    \hline
    \multirow{3}{*}{\shortstack[l]{\abg \\\textsl{\add,\bpemb,\gpt Mix}}}&Precision & $0.41 \pm 0.04$ & $0.32 \pm 0.03$ & $0.28 \pm 0.06$ & $0.27 \pm 0.05$ \\
    &Recall & $0.50 \pm 0.04$ & $0.41 \pm 0.02$ & $0.46 \pm 0.05$ & $0.62 \pm 0.07$ \\
    &F1 (macro) & $0.72 \pm 0.01$ & $0.68 \pm 0.01$ & $0.66 \pm 0.02$ & $0.68 \pm 0.02$ \\
    \hline
    \end{tabular}
    \caption{Comparison of augmentation techniques for 20x augmentation on \seed/\cls{identity-hate}: 
    means for precision, recall and macro-averaged F1-score shown with standard deviations (10~repetitions). 
    Precision and recall for \cls{identity-hate}; F1-score macro-averaged from both classes.
}
    \label{tab:results-identity}
  \end{center}
\end{table*}

\section{Augmentation computation performance}
\label{sec:augmentation_resources}


Table~\ref{tab:augmentation-resources} reports computational resources required for replicating augmentations. GPU computations were performed on a GeForce RTX 2080 Ti.
CPU computations were performed with an Intel Core i9-9900K CPU @ 3.60GHz with 8 cores, where applicable. 
Memory usage was collected using \textsl{nvidia-smi} and \textsl{htop} routines. Usage is rounded to nearest 100 MiB. 
Computation time includes time to load library from file and is rounded to nearest integer. 
Computation time (training and prediction) is shown separately for \gpt. 

\begin{table}[tbh]
  \begin{center}
    \begin{tabular}{lcccc}
    \hline    
    & \multicolumn{4}{c}{Augmentation} \\
    & \multicolumn{2}{c}{Memory (MiB)} & \multicolumn{2}{c}{Runtime (s)} \\
    \hline     
     & GPU &  CPU &  GPU & CPU \\
    \simple & - & - & - & $< 1$ \\
    \eda  & - & 100 & - & $1$ \\
    \add  & - & - & - & $1$ \\
    \wn  & - & 4000 & - & $1$ \\
    \ppdb & - & 2900 & - & $3$ \\
    \gensim & - & 600 & - & $32$ \\
    \bpemb & - & 100 & - & $<1$ \\
    \gpt & $3600$ & $3600$ & $12$ + $78$ & - \\
    \hline
    \end{tabular}
    \caption{Computational resources (MiB and seconds) required for augmenting 25 examples to 500 examples. \added[id=M]{\gpt takes approximately 6 seconds to train per epoch, and 3 seconds to generate 19 new documents.}}
    \label{tab:augmentation-resources}
  \end{center}
\end{table}

We provide library versions in Table~\ref{tab:augmentation-libraries}. 
We used sklearn.metrics.precision\_recall\_fscore\_support\footnote{\url{https://scikit-learn.org/stable/modules/generated/sklearn.metrics.roc_auc_score.html}} for calculating minority-class precision, recall and macro-averaged F1-score. For the first two, we applied \textit{pos\_label=1}, and set \textit{average = 'macro'} for the third.
For \rocauc, we used sklearn.metrics.roc\_auc\_score\footnote{\url{https://scikit-learn.org/stable/modules/generated/sklearn.metrics.roc_auc_score.html}} with default parameters. 
For t-tests, we used scipy.stats.ttest\_rel\footnote{\url{https://docs.scipy.org/doc/scipy/reference/generated/scipy.stats.ttest_rel.html}}, which gives p-values for two-tailed significance tests. We divided the p-values in half for the one-tailed significance tests.


\begin{table}[tbh]
  \begin{center}
    \begin{tabular}{lc}
    \hline    
    \textbf{Library} & \textbf{Version} \\ 
    \texttt{\small{https://github.com/}} & \multirow{2}{*}{\textit{Nov 8, 2019}\footnote{Latest version at time of submission.}} \\ \texttt{\small{jasonwei20/eda\_nlp}} &  \\%
    \texttt{apex} & 0.1 \\
    \texttt{bpemb} & 0.3.0 \\
    \texttt{fast-bert} & 1.6.5 \\
    \texttt{gensim} & 3.8.1 \\
    \texttt{nltk} & 3.4.5 \\ 
    \texttt{numpy} & 1.17.2 \\
    \texttt{pywsd} & 1.2.4 \\
    \texttt{scikit-learn} & 0.21.3 \\
    \texttt{scipy} & 1.4.1 \\
    \texttt{spacy} & 2.2.4 \\
    \texttt{torch} & 1.4.0\\ 
    \texttt{transformers} & 2.8.0 \\
    \hline
    \end{tabular}
    \caption{Library versions required for replicating this study. Date supplied if no version applicable. }
    \label{tab:augmentation-libraries}
  \end{center}
\end{table}

\section{Classifier training and testing performance}
\label{app:clf}

Table~\ref{tab:classifier-resources} specifies the system resources training and prediction required on our setup (Section~\ref{sec:augmentation_resources}). The \seed dataset has 8,955 documents and test dataset 63,978 documents. 
We used the 12-layer, 768-hidden, 12-heads, 110M parameter BERT-Base, Uncased-model.\footnote{\url{https://storage.googleapis.com/bert_models/2018_10_18/uncased_L-12_H-768_A-12.zip}}

%



\begin{table}[tbh]
  \begin{center}
    \begin{tabular}{lcccc}
    \hline    
    & \multicolumn{4}{c}{Training} \\
    & \multicolumn{2}{c}{Memory (MB)} & \multicolumn{2}{c}{Runtime (s)} \\
    \hline     
     & GPU &  CPU &  GPU & CPU \\
    \charlr & - & 100 & - & $4$ \\
    \wordlr & - & 100 & - & $3$ \\
    \cnn    & 400 & 400 & - & $13$ \\
    \bert   & 3800 & 1500 & 757 & - \\
    \hline
    \multicolumn{5}{c}{Prediction} \\
    & \multicolumn{2}{c}{Memory (MB)} & \multicolumn{2}{c}{Runtime (s)} \\
    \hline     
     & GPU &  CPU &  GPU & CPU \\
    \charlr & - & 100 & - & $25$ \\
    \wordlr & - & 100 & - & $5$ \\
    \cnn    & 400 & 400 & - & $42$ \\
    \bert   & 4600 & 4200 & $464$ & - \\
    \hline
    \end{tabular}
    \caption{Computational resources (MB and seconds) required for training classifiers on the \seed dataset and test dataset. Note that \bert~results here were calculated with mixed precision arithmetic (currently supported by Nvidia Turing architecture). 
    We measured memory usage close to 13 GB in the general case.}
    \label{tab:classifier-resources}
  \end{center}
\end{table}

\section{Lemma inflection in \wn}
\label{app:wordnet-inflection}

Lemmas appear as uninflected lemmas WordNet. To mitigate this limitation, we used a dictionary-based method for mapping lemmas to surface manifestations with NLTK part-of-speech (POS) tags. For deriving the dictionary, we used $8.5$ million short sentences ($\leq 20$ words) from seven corpora:
Stanford NMT,\footnote{\url{https://nlp.stanford.edu/projects/nmt/}}
OpenSubtitles 2018,\footnote{\url{http://opus.nlpl.eu/OpenSubtitles2018.php}}
Tatoeba,\footnote{\url{https://tatoeba.org}}
SNLI,\footnote{\url{https://nlp.stanford.edu/projects/snli/}}
SICK,\footnote{\url{http://clic.cimec.unitn.it/composes/sick.html}}
Aristo-mini (December 2016 release),\footnote{\url{https://www.kaggle.com/allenai/aristo-mini-corpus}}
and WordNet example sentences.\footnote{\url{http://www.nltk.org/_modules/nltk/corpus/reader/wordnet.html}}
The rationale for the corpus was to have a large vocabulary along with relatively simple grammatical structures, to maximize both coverage and the correctness of POS-tagging.
We mapped each lemma-POS-pair to its \myemph{most common inflected form} in the corpus.
When performing synonym replacement in \wn augmentation, we lemmatized and POS-tagged the original word with NLTK, chose a random synonym for it, and then inflected the synonym with the original POS-tag if it was present in the inflection dictionary.


\section{\gpt parameters}
\label{app:gpt2}

Table~\ref{tab:gpt} shows the hyperparameters we used for fine-tuning our \gpt models, and for generating outputs. 
Our fine-tuning follows the \texttt{transformers} examples with default parameters.\footnote{\url{https://github.com/huggingface/transformers/blob/master/examples/language-modeling/run_language_modeling.py}}

For generation, we trimmed input to be at most 100 characters long, further cutting off the input at the last full word or punctuation to ensure generated documents start with full words. Our generation script follows \texttt{transformers} examples.\footnote{\url{https://github.com/huggingface/transformers/blob/818463ee8eaf3a1cd5ddc2623789cbd7bb517d02/examples/run_generation.py}}

\begin{table}[h]
  \begin{center}
    \begin{tabular}{lc}
    \hline    
    \multicolumn{2}{c}{Fine-tuning} \\ 
    \hline    
    Batch size & 1 \\
    Learning rate & 2e-5 \\
    Epochs & 2 \\
    \hline
    \multicolumn{2}{c}{Generation} \\ 
    \hline    
    Input cutoff & 100 characters \\
    Temperature & 1.0 \\
    Top-p & 0.9 \\
    Repetition penalty & 1 \\
    Output cutoff & \multirow{2}{*}{\shortstack[l]{100 subwords or \\ EOS generated}} \\
     & \\
    \hline    
    \end{tabular}
    \caption{\gpt~parameters.}
    \label{tab:gpt}
  \end{center}
\end{table}

In \S4.2 -- \S4.4, we generated novel documents with \gpt fine-tuned on \cls{threat} documents in \seed for 2 epochs. 
In Table~\ref{tab:gpt-training-epochs}, we show the impact of changing the number of fine-tuning epochs for \gpt. Precision generally increased as the number of epochs was increased. However, recall simultaneously decreased. 

\begin{table*}[tbh]
  \begin{center}
    \begin{tabular}{llcccccccccc}
    \hline
    \multirow{2}{*}{\textbf{Classifier}}& \multirow{2}{*}{\textbf{Metric}}& \multicolumn{10}{c}{\textbf{Fine-tuning epochs on \gpt}} \\
     &&1&2&3&4&5&6&7&8&9&10\\
     \hline
    \multirow{3}{*}{\charlr}&Precision & 0.38 & 0.43 & 0.45 & 0.49 & 0.51 & 0.49 & 0.52 & 0.50 & \textbf{0.51} & \textbf{0.51} \\
    &Recall & \textbf{0.34} & \textbf{0.34} & 0.32 & 0.31 & 0.31 & 0.29 & 0.28 & 0.28 & 0.27 & 0.28 \\
    &F1 (macro) & 0.68 & \textbf{0.69} & 0.68 & 0.68 & \textbf{0.69} & 0.68 & 0.68 & 0.68 & 0.68 & 0.68 \\
    \hline
    \multirow{3}{*}{\wordlr}&Precision & 0.30 & 0.33 & 0.34 & 0.34 & \textbf{0.36} & 0.35 & 0.35 & 0.34 & 0.34 & 0.34 \\
    &Recall & \textbf{0.47} & 0.45 & 0.43 & 0.40 & 0.40 & 0.38 & 0.37 & 0.36 & 0.35 & 0.35 \\
    &F1 (macro) & 0.68 & \textbf{0.69} & \textbf{0.69} & 0.68 & 0.68 & 0.68 & 0.67 & 0.67 & 0.67 & 0.67 \\
    \hline
    \multirow{3}{*}{\cnn}&Precision & 0.26 & 0.28 & 0.30 & 0.32 & \textbf{0.33} & 0.32 & 0.31 & 0.31 & 0.31 & 0.32 \\
    &Recall & 0.49 & \textbf{0.50} & 0.47 & \textbf{0.50} & 0.48 & 0.48 & 0.48 & 0.46 & 0.47 & 0.46 \\
    &F1 (macro) & 0.66 & 0.67 & 0.68 & \textbf{0.69} & \textbf{0.69} & 0.68 & 0.68 & 0.68 & 0.68 & 0.68 \\
    \hline
    \multirow{3}{*}{\bert}&Precision & 0.11 & 0.14 & 0.15 & 0.15 & 0.16 & 0.17 & 0.17 & \textbf{0.19} & 0.17 & 0.17 \\
    &Recall & 0.62 & 0.66 & \textbf{0.67} & 0.64 & 0.65 & 0.62 & 0.62 & 0.62 & 0.61 & 0.61 \\
    &F1 (macro) & 0.59 & 0.61 & 0.62 & 0.62 & 0.62 & 0.63 & 0.63 & \textbf{0.64} & 0.63 & 0.62 \\
    \hline
    \end{tabular}
    \caption{Impact of changing number of fine-tuning epochs on \gpt-augmented datasets. Mean results for 10 repetitions. Highest numbers highlighted in bold. }
    \label{tab:gpt-training-epochs}
  \end{center}
\end{table*}

\section{Ablation study}
\label{app:ablation}

In \S4.2 -- \S4.4, we investigated several word replacement techniques with a fixed change rate. In those experiments, we allowed 25\% of possible replacements.
Here we study each augmentation technique's sensitivity to the replacement rate. As done in previous experiments, we ensured that at least one augmentation is always performed. Experiments are shown in tables~\ref{tab:ppdb-change-rate}--\ref{tab:bpemb-change-rate}. 

Interestingly, all word replacements decreased classification performance with \bert. We suspect this occurred because of the pre-trained weights in \bert.

We show \cls{threat} precision, recall and macro-averaged F1-scores for \ppdb in Table~\ref{tab:ppdb-change-rate}. 
Changing the substitution rate had very little impact to the performance on any classifier. 
This indicates that there were very few n-gram candidates that could be replaced. 
We show results on \wn in Table~\ref{tab:wn-change-rate}. 
As exemplified for substitution rate 25\% in~\ref{app:examples}, \ppdb and \wn substitutions replaced very few words. 
Both results were close to \simple (\S4.2, Table~6). 

We show results for \gensim in Table~\ref{tab:gensim-change-rate}. 
\wordlr performed better with higher substitution rates (increased recall). 
Interestingly, \charlr performance (particularly precision) dropped with \gensim compared to using \simple. 
For \cnn, smaller substitution rates seem preferable, since precision decreased quickly as the number of substitutions increased.

\bpemb results in Table~\ref{tab:bpemb-change-rate} are consistent across the classifiers \charlr, \wordlr and \cnn. 
Substitutions in the range 12\%--37\% increased recall over \simple. However, precision dropped at different points, depending on the classifier. 
\cnn precision dropped earlier than on other classifiers, already at 25\% change rate.

\section{Augmented \cls{threat} examples}
\label{app:examples}

We provide examples of augmented documents in Table~\ref{tab:example}. 
We picked a one-sentence document as the seed.
We remark that augmented documents created by \gpt have the highest novelty, 
but may not always be considered \cls{threat} (see example \gpt \#1. in Table~\ref{tab:example}). 

\begin{table}[!h]
  \begin{center}
    \begin{tabular}{lcccccc}
    \hline
    \multirow{2}{*}{\textbf{Metric}}& \multicolumn{6}{c}{\textbf{\ppdb: N-gram substitution rate}} \\
     &0&12&25&37&50&100 \\ 
    \hline
    & \multicolumn{6}{c}{\textbf{\charlr}} \\
    \hline
    Pre. & \textbf{0.14} & \textbf{0.14} & 0.13 & 0.13 & 0.13 & \textbf{0.14} \\
    Rec. & \textbf{0.09} & \textbf{0.09} & 0.09 & 0.08 & 0.07 & 0.05 \\
    F1 ma. & \textbf{0.55} & \textbf{0.55} & \textbf{0.55} & \textbf{0.55} & 0.54 & 0.54 \\
    \hline
    & \multicolumn{6}{c}{\textbf{\wordlr}} \\    
    \hline
    Pre. & 0.32 & 0.33 & 0.38 & \textbf{0.44} & 0.41 & 0.34 \\
    Rec. & \textbf{0.04} & \textbf{0.04} & \textbf{0.04} & \textbf{0.04} & 0.03 & 0.01 \\
    F1 ma. & \textbf{0.53} & \textbf{0.53} & \textbf{0.53} & \textbf{0.53} & \textbf{0.53} & 0.51 \\
    
    \hline
    & \multicolumn{6}{c}{\textbf{\cnn}} \\    
    \hline
    Pre. & \textbf{0.44} & 0.41 & 0.39 & 0.36 & 0.38 & 0.32 \\
    Rec. & 0.09 & 0.09 & \textbf{0.10} & 0.09 & 0.08 & 0.05 \\
    F1 ma. & \textbf{0.57} & \textbf{0.57} & \textbf{0.57} & \textbf{0.57} & 0.56 & 0.54 \\
    \hline
    & \multicolumn{6}{c}{\textbf{\bert}} \\
    \hline
    Pre. & 0.45 & 0.45 & 0.46 & 0.46 & 0.47 & \textbf{0.48} \\
    Rec. & \textbf{0.37} & \textbf{0.37} & \textbf{0.37} & 0.35 & 0.33 & 0.25 \\
    F1 ma. & \textbf{0.70} & \textbf{0.70} & \textbf{0.70} & \textbf{0.70} & 0.69 & 0.66 \\
    \hline
    \end{tabular}
    \caption{Impact of changing the proportion of substituted words on \ppdb-augmented datasets. Mean results for 10 repetitions. Classifier's highest numbers highlighted in bold.}
    \label{tab:ppdb-change-rate}
  \end{center}
\end{table}    

\begin{table}[!h]
  \begin{center}
    \begin{tabular}{llcccccc}
    \hline
    \multirow{2}{*}{\textbf{Metric}}& \multicolumn{6}{c}{\textbf{\wn: Word substitution rate}} \\
     &0&12&25&37&50&100 \\ 
    \hline
    & \multicolumn{6}{c}{\textbf{\charlr}} \\
    \hline
    Pre. & \textbf{0.15} & \textbf{0.15} & 0.14 & 0.14 & 0.12 & 0.10 \\
    Rec. & \textbf{0.10} &\textbf{ 0.10} & \textbf{0.10} & \textbf{0.10} & 0.09 & 0.07 \\
    F1 ma. & \textbf{0.56} & \textbf{0.56} & \textbf{0.56} & 0.55 & 0.55 & 0.54 \\
    \hline
    & \multicolumn{6}{c}{\textbf{\wordlr}} \\    
    \hline
    Pre. & 0.28 & 0.29 & 0.30 & 0.31 & \textbf{0.34} & 0.31 \\
    Rec. & 0.04 & 0.04 & 0.04 & \textbf{0.05} & 0.04 & 0.02 \\
    F1 ma. &  0.53 & 0.53 & 0.53 &\textbf{0.54} & \textbf{0.54} & 0.52 \\
    \hline
    & \multicolumn{6}{c}{\textbf{\cnn}} \\    
    \hline    
    Pre. & 0.42 & 0.43 & 0.42 & \textbf{0.45} & 0.44 & 0.32 \\
    Rec. & 0.10 & 0.11 & 0.11 & \textbf{0.12} & 0.10 & 0.07 \\
    F1 ma. & 0.58 & 0.58 & 0.58 & \textbf{0.59} & 0.58 & 0.55 \\
    \hline
    & \multicolumn{6}{c}{\textbf{\bert}} \\    
    \hline    
    Pre. & \textbf{0.45} & 0.44 & 0.43 & 0.43 & 0.42 & 0.35 \\
    Rec. & \textbf{0.31} & \textbf{0.31} & 0.29 & 0.26 & 0.24 & 0.18 \\
    F1 ma. & \textbf{0.68} & \textbf{0.68} & 0.67 & 0.66 & 0.65 & 0.61 \\
    \hline    
    \end{tabular}
    \caption{Impact of changing the proportion of substituted words on \wn-augmented datasets. Mean results for 10 repetitions. Classifier's highest numbers highlighted in bold.}
    \label{tab:wn-change-rate}
  \end{center}
\end{table}    

\begin{table}[!h]
  \begin{center}
    \begin{tabular}{llcccccc}
    \hline
    \multirow{2}{*}{\textbf{Metric}}& \multicolumn{6}{c}{\textbf{\gensim: Word substitution rate}} \\
     &0&12&25&37&50&100 \\ 
    \hline
    & \multicolumn{6}{c}{\textbf{\charlr}} \\    
    \hline     
    Pre. & 0.16 & 0.15 & 0.14 & 0.14 & 0.14 & \textbf{0.32} \\
    Rec. & 0.11 & 0.12 & \textbf{0.13} & \textbf{0.13} & \textbf{0.13} & 0.05 \\
    F1 ma. & 0.56 & 0.56 & \textbf{0.57} & \textbf{0.57} &\textbf{ 0.57} & 0.54 \\
    \hline
    & \multicolumn{6}{c}{\textbf{\wordlr}} \\    
    \hline     
    Pre. & 0.31 & \textbf{0.37} & 0.35 & 0.33 & 0.33 & 0.30 \\
    Rec. & 0.07 & 0.10 & 0.16 & \textbf{0.19} & \textbf{0.19} & 0.09 \\
    F1 ma. & 0.55 & 0.58 & 0.61 & \textbf{0.62} & \textbf{0.62} & 0.57 \\
    \hline
    & \multicolumn{6}{c}{\textbf{\cnn}} \\    
    \hline     
    Pre. & 0.41 & \textbf{0.44} & 0.39 & 0.35 & 0.28 & 0.15 \\
    Rec. & 0.13 & 0.18 & 0.19 & \textbf{0.20} & 0.17 & 0.06 \\
    F1 ma. & 0.59 & \textbf{0.62} & \textbf{0.62} & \textbf{0.62} & 0.60 & 0.54 \\
    \hline
    & \multicolumn{6}{c}{\textbf{\bert}} \\    
    \hline     
    Pre. & \textbf{0.44} & 0.43 & 0.40 & 0.36 & 0.33 & 0.13 \\
    Rec. & \textbf{0.35} & 0.27 & 0.16 & 0.13 & 0.11 & 0.03 \\
    F1 ma. & \textbf{0.69} & 0.66 & 0.61 & 0.59 & 0.58 & 0.52 \\
    \hline
    \end{tabular}
    \caption{Impact of changing the proportion of substituted words on \gensim-augmented datasets. Mean results for 10 repetitions. Classifier's highest numbers highlighted in bold.}
    \label{tab:gensim-change-rate}
  \end{center}
\end{table}    

\begin{table}[!h]
  \begin{center}
    \begin{tabular}{llcccccc}
    \hline
    \multirow{2}{*}{\textbf{Metric}}& \multicolumn{6}{c}{\textbf{\bpemb: Subword substitution rate}} \\
     &0&12&25&37&50&100 \\ 
    \hline
    & \multicolumn{6}{c}{\textbf{\charlr}} \\    
    \hline     
    Pre. & \textbf{0.65} & 0.64 & 0.56 & 0.52 & 0.49 & 0.37 \\
    Rec. & 0.17 & 0.20 & \textbf{0.22} & 0.20 & 0.17 & 0.06 \\
    F1 ma. & 0.63 & \textbf{0.65} & \textbf{0.65} & 0.64 & 0.63 & 0.55 \\
    \hline
    & \multicolumn{6}{c}{\textbf{\wordlr}} \\    
    \hline     
    Pre. & 0.26 & \textbf{0.34} & 0.31 & 0.30 & 0.25 & 0.19 \\
    Rec. & 0.07 & 0.13 & 0.22 & \textbf{0.25} & 0.23 & 0.13 \\
    F1 ma. & 0.55 & 0.59 & \textbf{0.63} & \textbf{0.63} & 0.62 & 0.57 \\
    \hline
    & \multicolumn{6}{c}{\textbf{\cnn}} \\    
    \hline     
    Pre. & \textbf{0.42} & 0.37 & 0.22 & 0.14 & 0.09 & 0.03 \\
    Rec. & 0.17 & 0.31 & \textbf{0.38} & 0.31 & 0.27 & 0.10 \\
    F1 ma. & 0.62 & \textbf{0.66} & 0.63 & 0.59 & 0.56 & 0.52 \\
    \hline
    & \multicolumn{6}{c}{\textbf{\bert}} \\    
    \hline      
    Pre. & \textbf{0.43} & 0.41 & 0.33 & 0.32 & 0.25 & 0.08 \\
    Rec. & \textbf{0.37} & 0.22 & 0.15 & 0.13 & 0.10 & 0.03 \\
    F1 ma. & \textbf{0.70} & 0.64 & 0.60 & 0.59 & 0.57 & 0.52 \\
    \hline
    \end{tabular}
    \caption{Impact of changing the proportion of substituted subwords on \bpemb-augmented datasets. Mean results for 10 repetitions. Classifier's highest numbers highlighted in bold.}
    \label{tab:bpemb-change-rate}
  \end{center}
\end{table}    

\begin{table*}[tbh]
  \begin{minipage}{16cm}
    \begin{tabular}{ll}
    \hline
    \# & Document sample \\
    \hline
    & \multicolumn{1}{l}{\seed: \seedL}    \\    
        & if you do not stop, the wikapidea nijas will come to your house and kill you \\
    \hline
    & \multicolumn{1}{l}{\simple: \simpleL}    \\
        1.& if you do not stop, the wikapidea nijas will come to your house and kill you \\
        2.& if you do not stop, the wikapidea nijas will come to your house and kill you \\
        3.& if you do not stop, the wikapidea nijas will come to your house and kill you \\  
    \hline
    & \multicolumn{1}{l}{\eda: \edaL\footnote[16]{\url{https://github.com/jasonwei20/eda_nlp}}}    \\
        1.& if you do \textit{put up} not stop the wikapidea nijas will come to your house and kill you \\
        2.& if you do not \textit{stopover} the wikapidea nijas will come to your house and kill you \\
        3.& if you do not \textit{break} the wikapidea nijas will come to your house and kill you \\
    \hline
    & \multicolumn{1}{l}{\add: \addL}    \\
        1.& \textit{We thank you both for your contributions to Wikipedia at-large and your use of the tool.} if you \\ 
        & do not stop, the wikapidea nijas will come to your house and kill you \\
        2.& if you do not stop, the wikapidea nijas will come to your house and kill you \textit{Honest! ))} \\
        3.& \textit{*** username, I am on a shared IP address.} if you do not stop, the wikapidea nijas will come \\
        &  to your house and kill you \\
    \hline
    & \multicolumn{1}{l}{\ppdb \ppdbL}    \\
        1.& if you do not \textit{be halted} , the wikapidea nijas will come to your \textit{home} and kill you \\
        2.& if you do not stop , the wikapidea nijas \textit{comes along} to your house and \textit{been murdered} you \\ 
        3.& if you do not stop , the wikapidea nijas will \textit{arrive} to your \textit{home} and kill you \\
    \hline
    & 
    \wn \wnL \\
        1.& if you do not stop , the wikapidea nijas will \textit{amount} to your house and kill you \\
        2.& if you do not stop , the wikapidea nijas will come to your house and \textit{down} you \\
        3.& if you do not stop , the wikapidea nijas will come to your \textit{theatre} and kill you \\
    \hline
    & \multicolumn{1}{l}{\gensim \gensimL}    \\
        1.& if you \textit{so} not stop , the wikapidea nijas will come to your\textit{s} house \textit{have} kill you \\
        2.& \textit{mean know} do not stop \textit{?} the wikapidea nijas will \textit{cos} to your house and kill you \\
        3.& if you do not \textit{thinking} , \textit{and} wikapidea nijas will come to your \textit{going} and kill you \\
    \hline
& \multicolumn{1}{l}{  \bpemb \bpembL }    \\   
    1.& \textit{whether} you \textit{know} not stop, the\textit{ldi}kapide\textit{ina}  ni\textit{ji} \textit{give} come to your house and kill you \\
    2.& \textit{shouldnever} do not stop, the wika\textit{myla} ni\textit{jag} will come to your house and kill \textit{want} \\
    3.& if you \textit{let} not \textit{train}, the \textit{w}ka\textit{na}pide\textit{i} nijas will come to\textit{never} house and \textit{dead} you \\
    \hline
%
    & \multicolumn{1}{l}{\gpt \gptL}    \\
    1.& \textit{for what you have done, you don't have a choice. You are too big and weak to stop.} \\
    2.& \textit{! So if you don't stop this, you will be murdered and your people will be ripped apart. I hope} \\
    & \textit{you die and die and die. No, you will die and die and die. Just stop and kill yourself for it.} \\
    & \textit{Advertisements} \\
    3.& \textit{. You will be hanged like a rat. The next day I will get a letter telling you that you were} \\
    & \textit{kicked out of your home by the wikapidea nijas and that you will die there. You are an idiot} \\ 
    & \textit{for calling the wikapidea nijas to death and you should die.} \\
    \hline
    & \multicolumn{1}{l}{
    \abg: \abgL
    } \\
    1.& \textit{Support - The other article is minor, and not worth a separate entry. }if you do not stop, the \\
    &  wikapidea nijas will come to your house and kill you \\
    2.& \textit{. You will be hanged like a rat. The next day I will get a letter telling you that you were} \\
    & \textit{kicked out of your home by the wikapidea nijas and that you will die there. You are an idiot} \\ 
    & \textit{for calling the wikapidea nijas to death and you should die.} \\
    3.& if you \textit{let} not \textit{train}, the \textit{w}ka\textit{na}pide\textit{i} nijas will come to\textit{never} house and \textit{dead} you \\
    \hline    
    \end{tabular}
    \caption{Documents generated by selected augmentation techniques in this study. Changes to original seed \textit{highlighted}. 
    The selected  sample is shorter than average (see \S3.1, Table~1).
    We anonymized the username in \add (\#3.). Three samples generated by each technique shown. }   
    \label{tab:example}
  \end{minipage}
\end{table*}

\end{document}